\documentclass[twocolumn]{article}

\usepackage{epsfig}
\usepackage{graphicx}
\usepackage{amsmath}
\usepackage{amssymb}

\usepackage[a4paper, top=20mm, bottom=20mm, left=20mm, right=20mm]{geometry}

\title{The Cross-Depiction Problem: Computer Vision Algorithms for Recognising Objects in Artwork and in Photographs}

\author{Hongping Cai\\
Department of Computer Science\\
University of Bath, UK\\
{\tt\small H.Cai@bath.ac.uk}
\and
Qi Wu\\
School of Computer Science\\
University of Adelaide, Australia\\
{\tt\small qi.wu01@adelaide.edu.au}
\and
Tadeo Corradi\\
Department of Computer Science\\
University of Bath, UK\\
{\tt\small ma1tmc@bath.ac.uk}
\and
Peter Hall\\
Department of Computer Science\\
University of Bath, UK\\
{\tt\small pmh@bath.ac.uk}
}

\begin{document}

\maketitle

\begin{abstract}

{\emph
The {\em cross-depiction problem} is that of recognising visual objects regardless of whether they are photographed, painted, drawn, {\em etc}. It is a potentially significant yet under-researched problem. Emulating the remarkable human ability to recognise objects in an astonishingly wide variety of {\em depictive forms} is likely to advance both the foundations and the applications of Computer Vision.

In this paper we benchmark classification, domain adaptation, and deep learning methods; demonstrating that none perform consistently well in the cross-depiction problem. Given the current interest in deep learning, the fact such methods exhibit the same behaviour as all but one other method: they show a significant fall in performance over inhomogeneous databases compared to their peak performance, which is always over data comprising photographs only. Rather, we find the methods that have strong models of spatial relations between parts tend to be more robust and therefore conclude that such information is important in modelling object classes regardless of appearance details.}

\end{abstract}

\section{Introduction}

Humans are able to recognise objects in an astonishing variety of forms. Whether photographed, drawn, painted, carved in wood, people can recognise horses, elephants, people, {\em etc}. The same is not true of computers,  even the very best recognition algorithms -- including deep learning -- exhibit a significant drop in performance when presented with an inhomogeneous data set, and fall further still when trying to recognise a drawn object after being trained only on photographic examples.

Cross-depiction forces one to consider which visual attributes  are necessary for recognition, and which are merely sufficient. To illustrate this: humans can recognise trains in full colour photographs, as vague fuzzy blobs in paintings such as
{\em Rain, Steam, and Speed} by J.M.W. Turner, in sketchy line drawings, as a simplified silhouette in UK road signs. Ostensibly at least, these vastly different depictions of a train have nothing in common -- except (of course) each of them shows a recognisable train.

\begin{figure}
\centerline{
\includegraphics[width=0.5\textwidth]{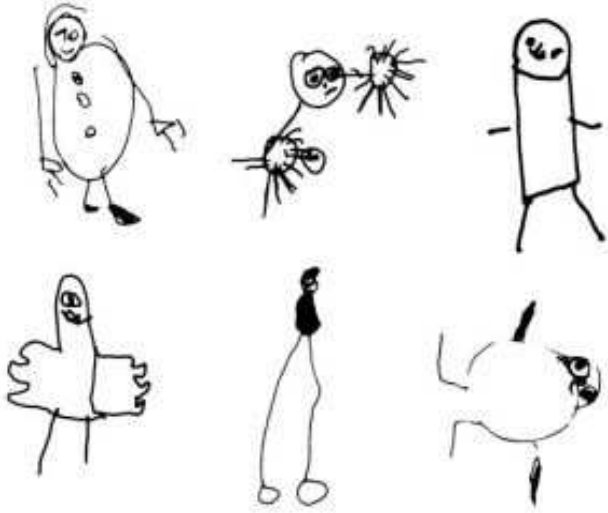}
}
\caption{Childrens' drawings.}
\label{fig:child}
\end{figure}

It is clear that specific appearance is able to vary significantly -- to a much greater degree than due to lighting changes, for example -- and still people can recognise objects.  Childrens' drawings, as in Figure~\ref{fig:child}, are both highly abstract and highly variable, yet contain sufficient information for objects to be recognised by humans, but not computers.

Equally clearly, learning the specifics of each depiction is at best unappealing, not least because the gamut of possible depictions is potentially unlimited. Rather, the question is {\em what abstraction do these classes have in common that allow then to be recognised regardless of depiction?} It is this and similar questions that push at the foundations of Computer Vision.

A machine that is able to recognise regardless of depiction  would provide a significant boost to current applications, such as image search and rendering. For example, given a photograph of the Queen of England, a search should return all portraits of her, including postage stamps that capture her likeness in bas-relief. Searching heterogeneous data sets is a real problem for the creative industries, because they archive vast quantities of material in a huge variety of depictions -- a problem that requires visual class models that spans depictive styles. Non-photorealistic rendering from images and video would be boosted too, not least because highly aesthetic renderings depend critically on the level of abstraction available to algorithms. Picture making is nothing like tracing over photographs: humans draw what they know of an object, not what they see -- computers should do like wise.

This paper: (1) establishes that there is a literature gap; (2) provides two databases designed for the cross depiction problem; (3) provides experimental evidence that no current method copes with the cross-depiction problem for either classification or detection; (4) provides an empirically based  explanation of the experimental results; (5) suggests possible ways ahead based on all the data in this paper.

As a note: in this paper, we use the term {\em photograph} as a short hand for ``natural image'', and the term {\em artwork} to refer to all other images.

\section{Related Literature}
\label{sec:lit}

There is a vast literature in Computer Vision that addresses the problem of {\em recognition}, by which we mean both {\em classification} (does this image contain an object of class X, or not?) and {\em detection} (an object of class X is at this place in this image). Yet almost no prior art addresses the cross-depiction, which is surprising given its genuine potential for advancing Computer Vision both in its foundations and in its applications.

Of the many approaches to visual object classification, the bag-of-words (BoW) family~\cite{csurka-eccv04,lazebnik-cvpr06,russakovsky-eccv12}
is amongst the most widespread. It models visual object classes as histograms of visual words; these words being clusters in feature space. Although the BoW methods address many difficult issues, they tend to generalise poorly across depictive styles (see Subsection~\ref{subsec:classresults}).  Alternative low-level features such as edgelets~\cite{gong-cvpr12,vlfeat} may be considered, or mid-level features such as region shapes ~\cite{jia2010classifying,gu2009recognition}. These features offer a little more robustness, but only if the silhouette shape is constrained -- and only if the picture offers discernible edges, which is not the case for many artistic pictures (Turner's paintings, for example).

Deformable models of various types are widely used to model  object classes for detection tasks, including several kinds of deformable template models \cite{cootes2001active,coughlan2000efficient}
and a variety of part-based models~\cite{amit2007pop,crandall2005spatial,
felzenszwalb2005pictorial,
felzenszwalb2010object,
fergus2003object,fischler1973representation,
leibe2008robust}.
In the constellation models from \cite{fergus2003object}, parts are constrained to be in a sparse set of locations, and their geometric arrangement is captured by a Gaussian distribution. In contrast, pictorial structure models \cite{felzenszwalb2005pictorial,felzenszwalb2010object,fischler1973representation} define a matching problem where parts have an individual match cost in a dense set of locations, and their geometric arrangement is captured by a set of spring connecting pairs of parts.  In those methods, the Deformable Part-based Model (DPM)  ~\cite{felzenszwalb2010object}, is widely used. It describes an object detection system based on mixtures of multi-scale deformable part models plus a root model. By modelling objects from different views with distinct models, it is able to detect large variations in pose. None of these directly address the cross-depiction problem.

Shape has also been considered.
Leordeanu {\em et al.} \cite{Leordeanu_etal_cvpr2007} encode relations between all pairs of edgels of shape to go beyond individual edgels. Similarly, Elidan {\em et al.} \cite{elidan2006learning} use pairwise spatial relations between landmark points. Ferrari et al. \cite{ferrari2008groups} propose a family of scale invariant local shape features formed by short chains of connected contour segments.
Shape skeletons are the dual of shape boundary, and also have been used as a descriptor. For example, Rom and Medioni \cite{rom1993hierarchical} suggest a hierarchical approach for shape description, combining local and global information, to obtain skeleton of shape. Sundar {\em et al.} \cite{sundar2003skeleton}  use skeletal graph to represent shape and use graph matching techniques to match and compare skeletons. Shock graph \cite{siddiqi1999shock} is derived from skeleton models of shapes, and focus on the properties of the surrounding shape. Shock graphs are obtained as a combination of singularities that arise during the evolution of a grassfire transform on any given shape. In particular, the set of singularities consists of corners, lines, bridges and other similar features. Shock graphs are then organised into shock trees to provide a rich description of the shape.

Algorithms usually assume that the training and test data are drawn from the same distribution. This assumption may be breached in real-world applications, leading to domain-adaptation methods such as transfer component analysis (TCA)~\cite{pan-nn11}, which transfer components from one domain to another. Both sampling geodesic flow (SGF) ~\cite{gopalan-iccv11} and geodesic flow kernel (GFK) ~\cite{gong-cvpr12} use intermediate subspaces on the geodesic flow connecting the source and target domain. GFK represents state-of-the-art performance on the standard cross-domain dataset \cite{fernando-iccv13}; it has been used to classify photographs acquired under different environmental conditions, at different times, or from different viewpoints. 

Cross-depiction problems are comparatively less well explored.
Some work is very specific -- Crowley and Zisserman take a weakly supervised approach, using a DPM to learn figurative art on Greek vases
~\cite{crowley2013gods}. Others develop the problem of searching a database of photographs based on a sketch query;  edge-based HoG was explored in \cite{hu-cviu13}, Li {\emph et al.}~\cite{li2013sketch}. Other have investigated sketch based retrieval of video ~\cite{hu2013markov,collomosse2009storyboard}.

\begin{figure*}[th!]
\begin{center}
\includegraphics[width=\textwidth]{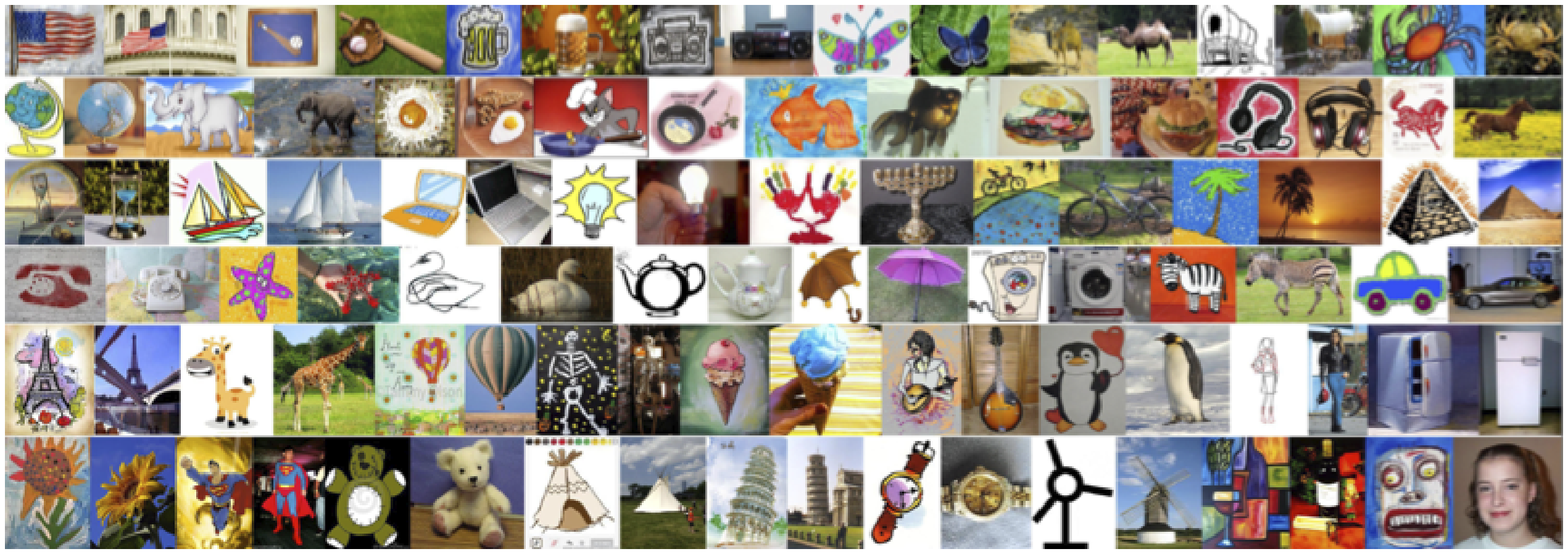}\\
\vspace*{6pt}
\includegraphics[width=\textwidth]{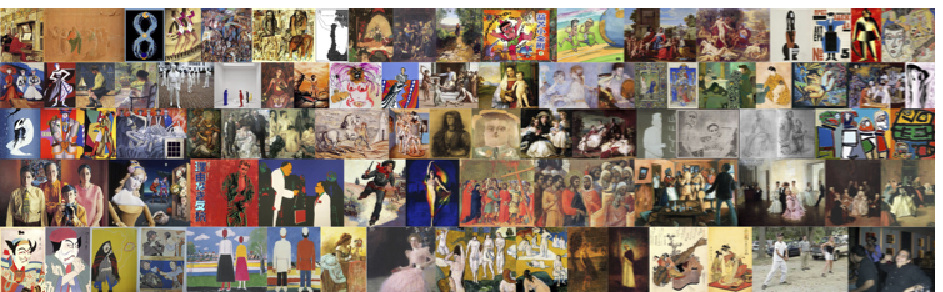}
\end{center}
   \caption{Top: Photo-Art-50 dataset: containing 50 object categories. Each category is displayed with one art image and one photo image. Bottom: People-Art databases, designed for detection.}
\label{fig:dataset}
\end{figure*}

Approaches to the more general cross depiction problem are rare. 
Matching visually similar images has been addressed using self similarity descriptors
~\cite{shechtman2007matching}.
It relies on a spatial map built from correlations of small patches; it therefore encodes a spatial distribution, but tends to be limited to small rigid objects. Crowley and Zisserman
~\cite{CrowleyZisserman_bmvc2014}
provide the only example of domain adaptation we know of specifically designed for the cross depiction problem; they train on photographs and then use midlevel patches to learn spatial consistencies (scale and translation) that allow matching from photographs into artwork. Their method performs well in retrieval tasks for 11 object classes in databases of paintings.

Classification, rather than matching, has also been studied. Shrivista {\em et al}~\cite{shrivastava-tog11} show that an Exemplar SVM trained on a huge database is capable of classification of both photographs and artwork. A less computationally intensive approach has been proposed ~\cite{wu-bmvc13} using a hierarchical graph model to obtain a coarse-to-fine arrangement of parts with nodes labelled by qualitative shape~\cite{wu2012prime}. Wu {\em et al} address the cross-depiction problem using a deformable model~\cite{wu2014learning}; they use a fully connected graph with learned weights on nodes (the importance of a nodes to discriminative classification), on edges (by analogy, the stiffness of a spring connecting parts), and multiple node labels (to account to different depictions); a method tested on 50 categories. Others use no labels at all, but rely on connection structure alone~\cite{xiao2011learning} or distances between low-level parts ~\cite{Leordeanu_etal_cvpr2007}.

Deep learning has recently emerged as a truly significant development in Computer Vision. It has been successful on conventional databases, and over a wide range of tasks, with recognition rates in excess of $90$\%. Deep learning has been used for the cross-depiction problem, but its success is less clear cut. Crowley and Zisserman~\cite{CrowleyZisserman_eccvvisart2014} are able to retrieve paintings in 10 classes at a success rate that does not rise above 55\%; their classes do not include people. Ginosar {\em et al}~\cite{Ginosar_etal_eccvvisart2014} use deep learning for detecting people in Picasso paintings, achieving rates of about 10\%.

Other than this paper, we know of only two studies assessing the performance of well established methods on the cross depiction problem. Crowley and Zisserman~\cite{CrowleyZisserman_bmvc2014} use a subset of the `Your Paintings' dataset~\cite{yourpaintings}, the subset decided by those that have been tagged with VOC categories~\cite{everingham2010pascal}. Using 11 classes, and objects that can only scale and translate, they report an overall drop in per class Prec@k (at $k=5$) from 0.98 when trained and tested on paintings alone, to 0.66 when trained on photographs and tested on paintings. Hu and Collomosse~\cite{hu-cviu13}  use  33 shape categories in Flickr to compare a range of descriptors SIFT, multi-resolution HOG, Self Similarity, Shape Context, Structure Tensor, and (their contribution) Gradient Field HOG. They  test a collection of 8 distance measures, reporting low mean average precision rates in all cases.


\section{Data Sets}
\label{sec:datasets}

To date there is no accepted publicly available database that has been specifically designed for the cross depiction problem. In this paper we use two annotated image datasets, both designed for the evaluation cross-depiction algorithms in classification and detection, samples can be seen in Figure~\ref{fig:dataset}, each is explained next.

\subsection{Multi-class Set: {\em Photo-Art-50}.}
\label{subsec:pa50}

This dataset is designed for classification problems. It contains 50 object classes, with  between 90 and 138 images for each class. Each class is approximately half photographs and half artwork. All 50 classes appear in Caltech-256; a few also appear in PASCAL VOC Challenge~\cite{everingham2010pascal} and ETH-Shape dataset~\cite{eth-shape}.

Some of the photographs come from Caltech-256, the rest from Google search. Arworks were searched using a few keywords to cover a wide gamut of depiction styles, \emph{e.g.}, `horse cartoon', `horse drawing', `horse painting', `horse sketches', `horse child drawing', \emph{etc}.  We manually selected images with a reasonable size of a meaningful object area. We further manually provide the ground-truth bounding boxes.

\subsection{Detection Set: {\em People-Art}}
\label{subsec:po}

However, one problem with building a dataset for cross-depiction is that objects classes do not appear with equal abundance in artwork. People tend to draw some object classes far more frequently than others -- people draw people a great deal, but artwork showing headphones and beer-mugs (both classes in Photo-Art-50, both in Caltech 256) is harder to come by, and (anecdotally/by observation) appear in relatively few depictive styles. Therefore our second database contains only people; it is better suited to detection problems than to classification problems.

It consists only of people, in 43 different styles, among which 41 styles are downloaded from {\tt wikipaintings.org} website, one cartoon style from  google search and one photographic style from PASCAL VOC2012. The dataset is divided into training set, validation set and test set. The training set has 1627 images, among which 1324 `person' objects are annotated in 521 images. The validation set has 1387 images, among which 1080 person objects are annotated in 442 images. The test set has 1617 images, among which 1083 person objects are annotated in 520 images.

This dataset represents a much wider gamut of depictive styles than Photo-Art-50. Additionally, the people in the artwork appear in far greater variety of poses than is common in photographs.

\section{Benchmarked Algorithms}
\label{sec:algs}

We benchmark several algorithms for classification and several for detection. Our purpose is not to act as advocates for any method, but to characterise current understanding with regard to the cross-depiction problem. 
The algorithms we report are not exhaustive: the area is far too well researched for that (at least using photographic databases). Rather, we have selected methods on the grounds of historical importance, current popularity, state of the art performance -- and have included some inventions of our own. In addition to the methods reported, we also tested other alternatives
\cite{shechtman2007matching}
\cite{eth-shape}
\cite{wu-bmvc13}
\cite{yang2011articulated}
but none worked sufficient well to report here.

\subsection{Bag of Words (BoW)}
\label{subsec:bow}

There are many variants of BoW methods, see Section~\ref{sec:lit}. We use Csurka {\em et al's} version~\cite{csurka-eccv04}; because it is well known, widely used, and classifies photographs well. Given a set of labelled training images, local descriptors are computed on a regular grid  with multiple-sized regions. A vocabulary of words is constructed by vector quantisation of local descriptors with k-means clustering ($k=1000$). To construct a visual class model (VCM) each image is partitioned into $L$ levels of increasingly fine cells ($L=2$ in our experiments). A histogram of word occurrences is built for each cell; concatenating these histograms encodes the image with a 5000 dimensional vector. A one-versus-all linear SVM classifier is trained on a $\chi^2$-homogeneous kernel map~\cite{vedaldi-cvpr10} of all training histograms. Given a test image the local features are extracted in the same way as in the training stage, mapped onto the codebook to build a multi-resolution histogram, which is then classified with the trained SVM.


Choice of feature may be important to the cross
depiction problem (see Section~\ref{sec:lit}).
Therefore we  test a collection of distinct features, as follows:

{\bf \em SIFT}~\cite{lowe-ijcv04} is a 128-dimensional vector created by stacking 8-bin orientation histograms on $4\times4$ cells. We use the implementation of dense-SIFT in \cite{vlfeat} and sample SIFT with four region sizes  on a regular grid with 3 pixels step.
{\bf \em Geometric Blur} (GB)~\cite{berg-cvpr01} describes local regions by geometrically blurring oriented edge maps. It is able to match object parts with very different appearance in two images.
We follow the original setup in~\cite{berg-cvpr01}.
{\bf \em Self-similarity desciptors} (SSD) \cite{shechtman2007matching, chatfield-iccv09} measure local self-similarity patterns by correlating a tiny local patch (typically $5 \times 5$) within a larger local region. It computes local correlations of patches
rather than pixel values, and performance well at matching similar objects invariant to depictive styles. We include it in the BoW framework to observe its behaviour in cross-depiction classification. We follow the default parameter settings from \cite{chatfield-iccv09} except that we use 4 region sizes to capture a wider variation of local patterns.
{\bf \em Histogram of Oriented Gradient} (HOG)~\cite{dalal-cvpr05} is a vector of normalised histograms from tiled block regions.
It is the most effective feature in the context of object detection~\cite{felzenszwalb-pami10}
and also the most favored local feature in the context of sketch-based retrieval~\cite{li2013sketch,eitz-tvcg11,eitz-siggraph12}. We compute HOG using the VLFeat~\cite{vlfeat} implementation.
The gradients are quantised into 9 orientations and four cell sizes are used.
{\bf \em edgeHOG} for comparison due to its effectiveness in sketch-based retrieval~\cite{hu-icip10}.
Unlike standard HOG which extracts the descriptor on the original image map, edgeHOG computes the gradient orientation histograms over edge maps.

{\bf \em Fisher Vectors (FV)}~\cite{perronnin-eccv10}, strictly speaking, are not BoW. Instead of counting the words occurrence (as in BoW). Given a set of local feature vectors (we use SIFT) extracted from training images, fitted a $K=256$-component GMM to their  distribution. The FV of an image is the stack of the mean and covariance deviation into a vector. We follow \cite{perronnin-eccv10}, applying Hellinger's kernel to each dimension of the Fisher vector followed by $L^2$-normalisation.
Like BoW, a spatial pyramid (identical) is used.  Then, a one-versus-all linear SVM classifier is trained on the Fisher vectors obtained from all training images


\subsection{Domain Adaptation}
\label{subsec:domadap}

Photographs and art can be seen as belonging to different domains.  Excellent domain adaptive methods, not limited to~\cite{gopalan-iccv11,gong-cvpr12,fernando-iccv13,saenko-eccv10,gong-icml13}, show clear benefits for photographs captured under different conditions. Two state of the art methods are evaluated on our dataset:

{\bf \emph{Geodesic Flow Kernel}}(GFK) models the source domain $\mathcal{S}$ and target domain $\mathcal{T}$ with lower dimensional linear subspaces and embeds them onto a Grassmann manifold. Geodesic flow is parameterized as a curve between these two subspaces, see Gong {\em et al}~\cite{gong-cvpr12}. We used two variants of GFK kernels: GFK\_PCA and GFK\_LDA. In GFK\_PCA the original features are projected onto the 49 dimensional subspace, with PCA on each domain. GFK\_LDA replaces PCA with a supervised dimension reduction method -- linear discriminant analysis (LDA) -- on the source domain.
{\bf \emph{Subspace Alignment}} (SA) \cite{fernando-iccv13} project $\mathcal{S}$ and $\mathcal{T}$ to respective subspaces. Then, a linear transformation function is learned to align the two spaces.


\subsection{Part Based Models}
\label{sec:dpm}

Part based models use larger scale `features', and take spatial relationship between these parts into account.

Deformable Parts Model {\bf \em  DPM}~\cite{felzenszwalb-pami10} is a state of the art representation. It models an object with a star graph, \emph{i.e.}, a root filter plus a set of parts. Given the location of the root and the relative location of $n$ parts; $n=8$ in our experiments. The score of the star model is the sum of responses of the root filter and parts filters, minus the displacement cost. Each node in a DPM is labelled with a HoG feature, learned from examples.

By analogy with domain adaptation, we considered the possibility  of query expansion for DPM to obtain Adapted DPM ({\bf ADPM)}.  We first train a standard DPM model for each object category in the training set (\emph{i.e.}, source domain) $\mathcal{S}$. We then apply the models on the test set (\emph{i.e.}, target domain) $\mathcal{T}$. A confidence set $\mathcal{C} \subset\mathcal{T}$ is constructed from the test set for training expansion by picking images that match a particular VCM especially well: 
\begin{equation}
\mathcal{C} = \{x \in \mathcal{T} | s_{1}(x) > \theta_1 
\wedge s_{1}(x)-s_{2}(x) > \theta_2\}
\end{equation}
with $s_1(x)$ the highest DPM score greater, and $s_2(x)$ the next highest score, and  $\theta_1,\theta_2$ are user-specified parameters to threshold the best score and margin respectively. We found  $\theta_1=-0.8$ and $\theta_2=0.1$ to be a good trade-off between minimising false positives (~5\%) and including appropriate number of expanded data (around 580 images in $\mathcal{C}$).

The fully connected multi-labelled graph {\bf \em MG} model of Wu {\em et al}~\cite{wu2014learning} is designed for the cross-depiction problem. It attempts to separate appearance features (contingent on the details of a particular depiction) from the information that characterises an object class without reference to any depiction. Unlike DPM, it comprises a fully connected weighted graph, and has multiple labels per node. Each graph has eight nodes. Weights on nodes can be interpreted as denoting the importance of a node to object class characterisation in a way that is independent of depiction. Weights on arcs are high if the distance between the connected pairs of parts varies little. These weights are learned using a structural support vector machine~\cite{Cho_etal_2013}. In addition to the weights, each node carries 2 features labels. These are designed to characterise the appearance of parts in both photographs and artwork (see the Discussion~\ref{sec:discuss} for a justification).

\begin{table*}[ht!]
\center
  \begin{tabular}{|c|c||c|c|c|c|c|c||c|c||c|}
  \hline
  \multicolumn{2}{|c||}{model} & \multicolumn{5}{c|}{BoW} & FV  & DPM  & MG  &  CNN
  \tabularnewline \hline
  train &  test &
  SIFT & GB & SSD & HOG & eHOG &
  SIFT & HOG & 2$\times$HOG  & learned
  \tabularnewline \hline 
  P  &  P & 
  84 & 77 & 66 & 72 & 70 & 87 &
  88 & 85 & 97
  \tabularnewline
  M  & P  & 
  80 & 72 & 58 & 65 & 63 & 84 &
  85 & 90 & 96
  \tabularnewline
  A  &  P & 
  64 & 60 & 39 & 42 & 50 & 66 &
  78 & 83 & 91
  \tabularnewline \hline 
  A  &  A & 
  74 & 72 & 49 & 55 & 60 & 77 &
  83 & 89 & 89
  \tabularnewline
  M  &  A & 
  69 & 67 & 45 & 50 & 56 & 73 &
  80 & 89 & 87
  \tabularnewline
  P  &  A & 
  44 & 50 & 31 & 29 & 40 & 47 &
  68 & 83 & 73 \tabularnewline
  \hline
  \end{tabular}
\vspace{2pt}
\caption{Classification Benchmarks on Photo-Art-50. Each row is a train (30 image) / test (rest) pattern: {\bf A}rt, {\bf P}hoto, {\bf M}ixed. Each column is an algorithm with feature, divided into groups: BoW~\protect{\cite{lowe-ijcv04,berg-cvpr01,shechtman2007matching,dalal-cvpr05,hu-icip10,perronnin-eccv10}}, parts-based~\protect{\cite{felzenszwalb-pami10,wu2014learning}}, deep learning with CNN~\protect{\cite{CrowleyZisserman_eccvvisart2014}}. Each cell shows the mean of 5 randomized trials. The standard deviation on any column never rises above 2\%.}
\label{tab:bow}
\end{table*}

\begin{table*}
\centerline{
\begin{tabular}{|c|c||c|c|c|c|c|c|c|}
\hline
train & test & SVM & PCA\_S & PCA\_T & GFK\_PCA & GFK\_LDA & SA & ADMP
\\
\hline
Art & Photo & 54 & 46 & 48 & 48 & 50 & 45 & 84\\
\hline
Photo & Art & 36 & 30 & 31 & 31 & 32 & 29 & 78\\
\hline
\end{tabular}
}
\vspace{2pt}
\caption{Domain Adaptation: These methods are designed to `jump' from one domain to the other, therefore we restricted training to one domain and testing on the other. Each column is a train/test pattern. Each column is an algorithm: SVM is Linear SVM using SIFT features; PCA\_S is SVM with PCA isn source domain only; PCA\_T is SVM with PCA on target only; GFK\_PCA and GGK\_LDA is GFK~\cite{gong-cvpr12} with PCA and LDA on feature; SA is subspace alignment
~\cite{fernando-iccv13}.}
\label{tab:domadap}
\end{table*}

\subsection{Deep Learning}
\label{subsec:DL}

Convolutional neural networks (CNN)~\cite{krizhevsky-nips12} has yielded a significant performance boost on image classification. To adapt the CNN to the object detection task, Girshick et al.~\cite{girshick-cvpr14} proposed {\bf \em R-CNN} (Regions with CNN features) by combining region proposals with CNNs. As the annotated data is scarce, it is insufficient to train a large CNN.

The solution we use is standard practice. The CNN parameters are first initialised by supervised pre-training from the large ILSVRC2013 dataset, then fine-tuned on the annotated regions. To be precise, the R-CNN method works in three steps. The first generates around 2000 category-independent region proposals by selective search~\cite{uijlings-ijcv13}. The second step is to extract a fixed-length feature vector for each wrapped region by forwarding it into a pre-trained AlexNet~\cite{krizhevsky-nips12}. More specifically, we use the output of the last fully-connected layer (fc-7) as the region features. The third step is class-specific linear SVM.

For classification, we follow Crowley and Zisserman~\cite{CrowleyZisserman_eccvvisart2014}, encoding images in each class with CNN features, which are then used  as input to learn a one-vs-all linear SVM classifier. For Detection, we run the experiments with R-CNN codes~\cite{girshick-cvpr14} downloaded from the authors website. The CNN architectures and the fine-tuning are implemented using the publicly available Caffe~\cite{jia-caffe14}.

\section{Classification Benchmarks}
\label{sec_classify}

We use Photo-Art-50 for classification benchmarking, with a variety of algorithms. We considered six different train/test patterns, given by the different combinations of training on photographs, art, or a mixed set; and testing on photos or art alone. In all cases we repeated the experiment 5 times,  randomly selecting 30 images for training, using the rest for testing.

For {\bf \em BoW} and {\bf \em FV}s: for each descriptor we built an SVM classifier using a $\chi^2$ kernel.
{\bf \em Domain adaptation} is about moving from one domain to another, so we only benchmarked photograph to art, and vice-versa. To monitor any affect of domain adaptation we built a control classifier: a linear SVM using SIFT features (SVM). We also using principle component analysis on both source (PCA\_S) and target (PCA\_T) domains individually to reveal the impact of PCA. We implemented geodesic kernel flow (GFK)~\cite{gong-cvpr12} with PCA and LDA applied to data, and also subspace alignment (SA)~\cite{fernando-iccv13} as high quality domain adaptation algorithms.
{\bf \em Deformable models} are used to classify by scanning an image in an effort to detect each class -- the class with the highest detection score is used as the class.
We follow~\cite{CrowleyZisserman_eccvvisart2014} who use CNN  for classifying art; we also include photographs.


\subsection{Results and Discussion for Classification}
\label{subsec:classresults}

Tables~\ref{tab:bow} and ~\ref{tab:domadap} show our results for the classification benchmarks.  Each row is a different train/test pattern, and each column a local feature descriptor. Each cell shows the percentage of correct classifications, averaged over 5 runs, rounded to the nearest integer.

General patterns emerge. It is clear the BoW and domain adaptation methods perform the least well. Models that take spatial relations into account perform better. What is most striking and surprising is that models using inter-part distance alone as a feature is comparable with R-CNN when photographs are used as the test set, and are the best performing of all when artwork is used as the test set (CLT excepted: classifier is important).

For most algorithms in Table~\ref{tab:bow}, training on photographs and testing on photograph yields the highest performance. For all algorithms in Table~\ref{tab:bow} training photographs and testing on art proves the most difficult case; training on art and testing on photographs is the most difficult case whenever the test set is restricted to photographs alone.

Against this data, the domain adaptation methods offer no advantage  -- with the exception of domain-adapted DPM. We note Crowley and Zisserman~\cite{CrowleyZisserman_bmvc2014} report a similar pattern of findings for their adaptation model (which includes spatial and feature adaptation).

Looking at details, the Fisher Vector is the best of the BoW-like method, which is  consistent with observations in \cite{perronnin-eccv10}. EdgeHOG outperforms the standard HOG when trained on artwork, which is consistent with the observations of \cite{eitz-tvcg11,hu-icip10}. Gaussian Blur kernels also capture edge information, which may explain why they drop away the least in the photo/art train/test pattern. The SSD performance is possibly surprising -- but results similar to our own have been observed by others interested in sketch-based classification task~\cite{hu-cviu13,li2013sketch}. Its poor performance it possibly explained by its need for relatively rigid objects.

\section{Detection Benchmarks}
\label{sec_det}

\begin{table}[t!]
\center
  \begin{tabular}{|c|c||c|c|c|}
  \hline
   train & test & DPM          & ADPM & MG  \tabularnewline
  \hline
  Photo & Photo & 96  & -- & 89 \tabularnewline
   Art  & Photo & 80  & 84  & 85 \tabularnewline
  \hline
    Art & Art   & 84  & -- & 88 \tabularnewline
   Photo &  Art  & 73  & 78  & 71 \tabularnewline
  \hline
  Photo + Art & Photo \& Art & 84 & -- & 89 \tabularnewline
  \hline
  \end{tabular}
  \vspace{2pt}
\caption{Detection Precision: Comparison of mean average precision (mAP) across photo-art domains on our dataset, 30 images per object for training. DPM and ADPM stand for DPM trained without and with cross-depiction expansion, respectively, MG is fully-connected multi-labelled graph.}
\label{tab:detectPA50}
\vspace{-5pt}
\center
  \begin{tabular}{|c||c|c|c|}
  \hline
  method & fine-tuning &  test & AP \tabularnewline
  \hline
   DPM  &   -     & art-test	& 32 \tabularnewline
 R-CNN  &   PASCAL VOC2012   & art-test	& 26 \tabularnewline
 R-CNN  &   art-trainval     & art-test & 40 \tabularnewline
  \hline
  \end{tabular}
\caption{Detection performances (average precision, AP) of DPM model and R-CNN model on Art-Person dataset.}
\label{tab:detPeople}
\end{table}


We carry out object detection on Photo-Art-50 for DPM,  adapted DPM (ADPM), and the fully-connected multi-labelled graph (MG).
To make a detection in an image, we first construct a dense multi-scale feature pyramid. Then we locate each node of the class model (DPM, ADPM, MG) at the best $k$ locations. We then build a structure used in matching, the exact scoring mechanism being determined by ~\cite{felzenszwalb-pami10} for DPM and ADPM and by ~\cite{wu2014learning} for MG. This is a deterministic algorithm, so was run exactly once per image. 

Detection rates on Photo-Art-50 are high, so we constructed the more challenging dataset, People-Art (Section~\ref{sec:datasets}). We detected people using DPM, R-CNN without domain refinement, and R-CNN with domain refinement. We see general purpose DPM outperforms R-CNN, unless the deep learning network is refined on artwork.


\subsection{Results and Discussion for Detection}

Table~\ref{tab:detectPA50} shows results using the Photo-Art-50 dataset. Each row shows a training / testing pattern, and each column an algorithm. Each cell is the mean average precision (mAP) across 50 objects, with a standard deviation of 2\%.
It shows that DPM performs very well when photographs are used for both training and testing; which is consistent with the previous work~\cite{felzenszwalb-pami10}. The fully connected multi-labeled graph (MG) outperforms DPM in all other cases except the case when photographs form the training set and artwork is used for testing; but  the standard deviation on the error is 2\%, so the difference is not significant.
. Echoing the classification task in Section~\ref{sec_classify} -- the performance of both DPM and MG drop significantly compared to other train/test patterns. The Adapted DPM shows to best performance on the heterogeneous train/test pattern, and does so beyond the 2\% deviation limit. This makes it somewhat significant.

Results for detections on the People-Art dataset are shown in Table~\ref{tab:detectPA50}. These show that DPM outperforms the R-CNN machine trained on photographs alone, but once R-CNN is tuned to the People-Art dataset it outperforms DPM.


\begin{figure}[t!]
\centering
\includegraphics[width=0.4\textwidth]{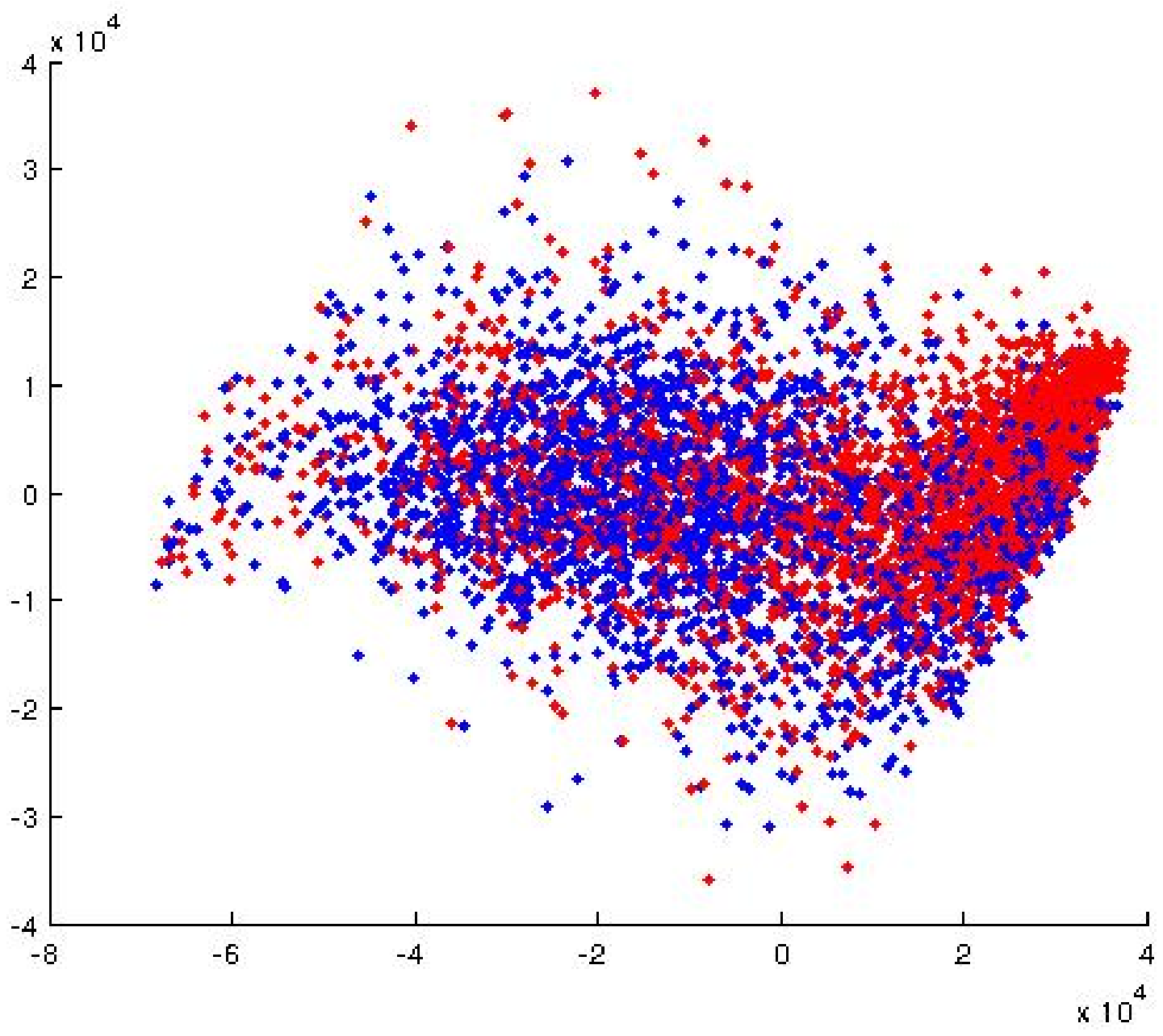}\\
\includegraphics[width=0.4\textwidth]{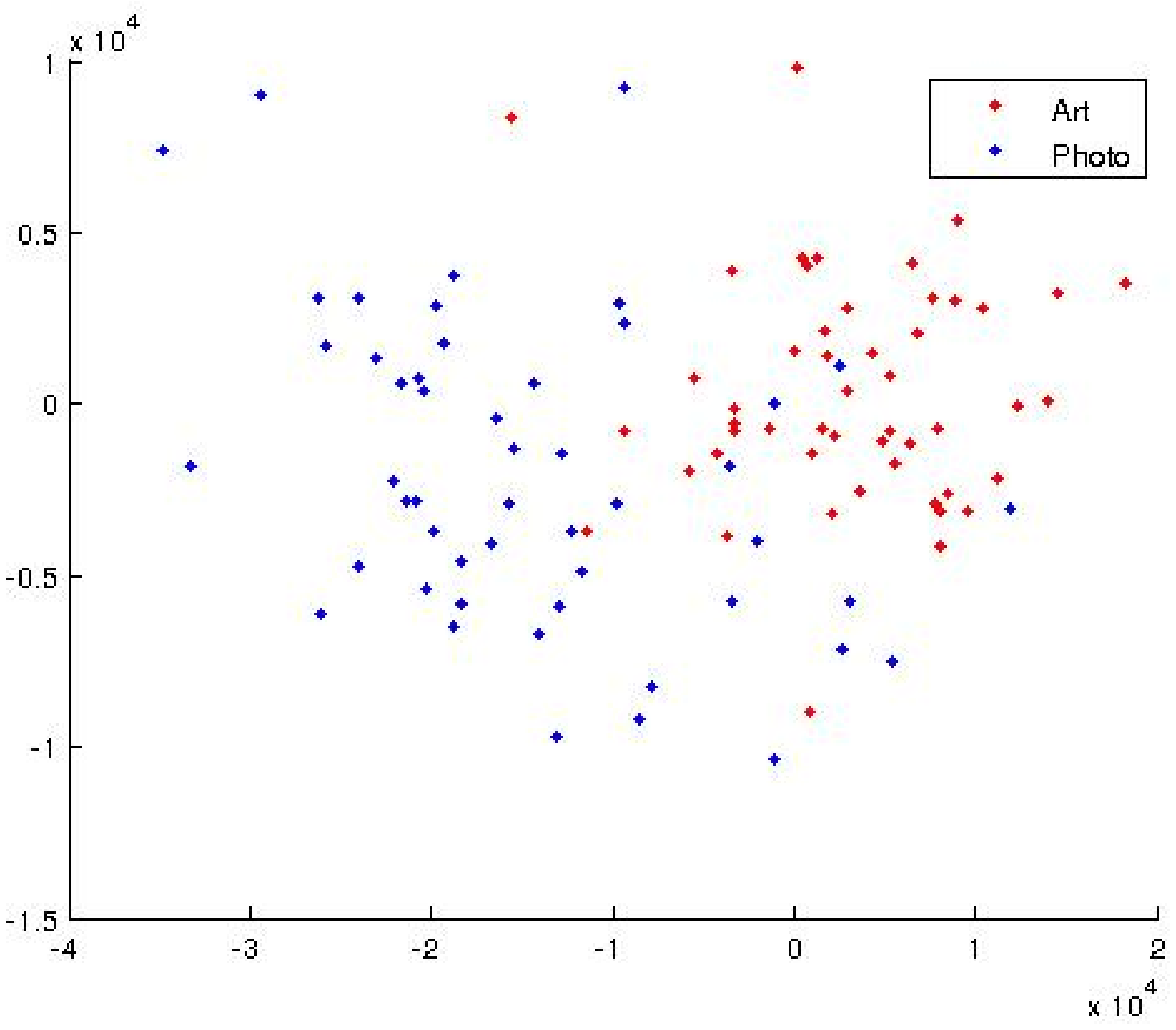}
\caption{Above: each image in Photo-Art-50 plotted in an eigenspace spanning raw images, art in red, photos in blue.
Below: The centre of each class in Photo-Art-50: red (art), blue(photo). The images and the cluster centres tend to form two groups: art/photo.}
\label{fig:centres}
\end{figure}

\section{General Discussion}
\label{sec:discuss}

Across all classification and detection experiments we the same trend: a  fall in performance in any case where art is included. This fall is most marked whenever photographs are used for training and artwork for testing, and is seen in all cases other than the M-Graph~\cite{wu2014learning}.

These observation need an explanation. Intuition suggests that the difference between the low-level images statistics of photographs and artwork differ is a cause. To investigate this we used all of the $10356$ raw image in Photo-Art-50 and rescaled the to square images with 256 pixels per edge. We then represented each image as a vector, and computed the covariance over all Art images and all photographs; the largest single eigenvalue for photograph  is $2.65 \times 10^{6}$, the largest single eigenvalue of artwork is $3.42 \times 10^3$. This shows that the variance over artworks is about 1000 times greater than the variance over photographs.

Results from a more detailed version of this experiment can be seen in Table~\ref{tab:kl}. There, the symmetric KL-divergence of different data sets that comprise two domains is computed. The domains $C$ (Caltech-256), $A$ (Amazon), $W$ (Webcam), and $D$ (DSLR) are all used in domain adaptation problems. The symmetric KL-divergence between these is shown along side the difference between art and photographs in Photo-Art-50. As can be seen, the distance between domains in Photo-Art-50 is by far the largest. This may explain the difficulty domain adaptation methods appear to face when `jumping the gap' -- in the cross-depiction problem, the gap is wider than the datasets usually used in domain adaptation.

A stronger hypothesis is this. Let $\mathbb{X}$ be an object class and $x_P \in \mathbb{X}$ be a photographic instance and $x_A$ is artwork instance of the that class. Similarly $y_P, y_A \in \mathbb{Y}$ are a photograph and artwork of class $Y$. Denote the set of all $x_p$ by $X_P$, meaning the `photo object', {\em etc}. Suppose too there is a measure $d(.,.)$. We expect the intra-class distance (same domain, different class) to be less than the inter-class distance for (different domain, same class). That is $d(x_P,x_A) > d(x_P,y_P)$, $d(x_P,x_A) > d(x_A,y_A)$, {\em etc.} ; which we call  `atomic hypothesis statements''.  To test this we used raw images Photo-Art-50 as raw input, each scaled to a square image of pixel width 256. We then mapped all the data in a 4 dimensional space using PCA over all the data, and computed an eigenmodel for each `photo object' $X_P$, and each `art object' $X_A$. We assumed a K-NN classifier, so that $X_P$ is represented by the mean, likewise $X_A$, and the measure is Euclidean distance. We found that a fraction $0.67$ of all possible atomic hypothesis statements to be true. This means the centres of objects in different classes are expected to be closer than the centres of the same object in different domains, which may confuse feature based classifiers and explain our results above.

These simple experiments help explain our benchmark results -- and results reported by others. They show that the distance between, artwork and photographs for any given class of object is expected to be larger than between photographs of different classes. That is, the variation due to depiction is greater than the variation due to object identity. This accounts for the reason training on photographs alone and testing on art gives the poorest results -- the photographs cluster into a relatively small region of feature space, and algorithms seem to over-fit the data in that small region. Artwork, on the other had, tends to be more varied than photographs, which is why training on Art and testing on Photographs tend to be a little more robust -- but training on a mixed set gives clearly the best results, because the training data span the full variance.

This wide variance in low-level statistics also helps explain the appeal of spatial information regarding object class identity. So far every method we have experimented that uses some kind of spatial information shows less fall away in the cross-depiction problem; this is true also of ~\cite{CrowleyZisserman_bmvc2014}. In this paper we see DPM outperform BoW, and the M-Graph outperform DPM. This result is in line with ({\em e.g.}) Leordeanu {\em et al.}~\cite{Leordeanu_etal_cvpr2007} who use the distance between low-level parts (edgelets) as a feature to characterise objects and achieve excellent detection results on the PASCAL dataset \cite{everingham2010pascal} of photographs; it may be effective too on Photo-Art-50, but this is to be proven.

\begin{figure}
\centerline{
\includegraphics[width=0.4\textwidth]{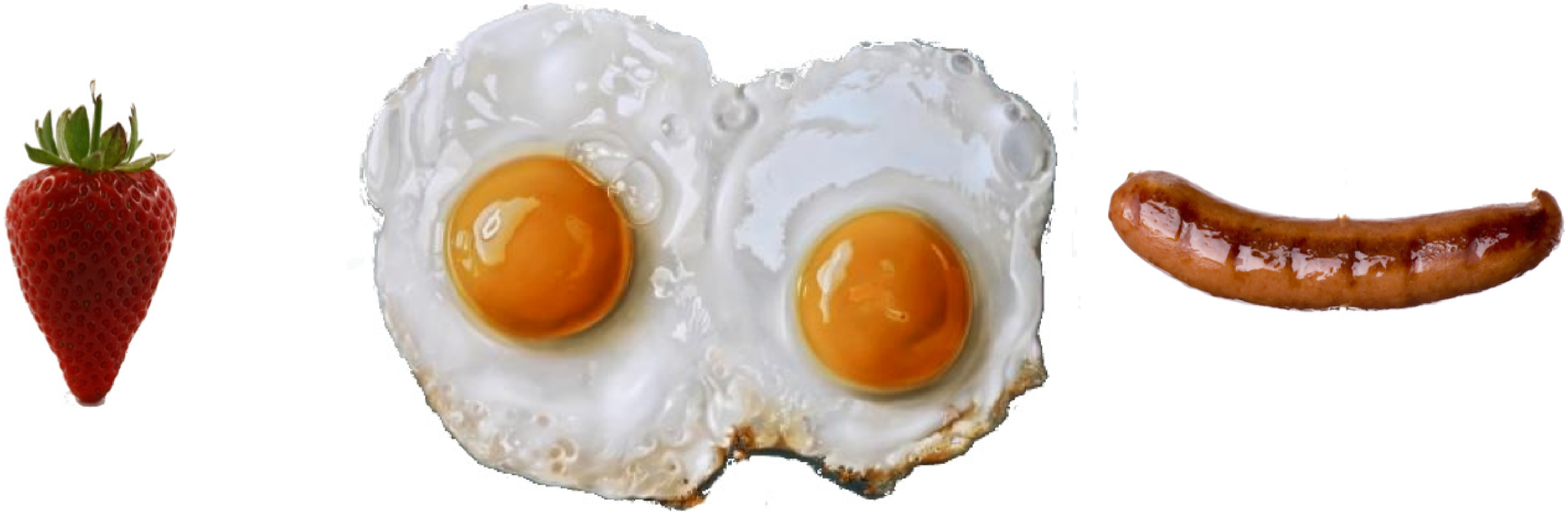}
}
\centerline{
\includegraphics[width=0.22\textwidth]{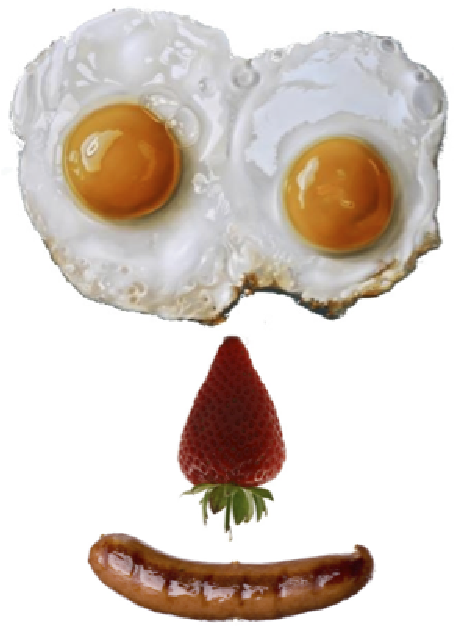}
}
\caption{The presence of a face depends on spatial arrangement of parts: above, no face; below smiling face.}
\label{fig:eggy}
\end{figure}

\begin{table}
\center
  \begin{tabular}{|c|c|c|c|c||c|}
  \hline
  \multicolumn{5}{|c||}{Cross-domain datasets~\cite{saenko-eccv10,gong-cvpr12}} & \emph{Photo-Art-50} \tabularnewline
  \hline
  \hline
C-A   &  C-D  &  A-W  & D-A   & D-W    & Photo-Art      \tabularnewline
  \hline
0.079 & 0.271 & 0.239 & 0.292 & 0.047 & \textbf{0.466}  \tabularnewline
  \hline
  \end{tabular}
\vspace{1pt}
\caption{Comparison of symmetric K-L divergence $\mathcal{D}(P_1,P_2)$ between domain pairs. Four domain sets in~\protect{\cite{saenko-eccv10,gong-cvpr12}}: C - Caltech-256, A - Amazon, W - WebCam, D - DSLR.}
\label{tab:kl}
\end{table}

This empirical data is supported anecdotally. The childrens' drawings in Figure~\ref{fig:child} are clearly people, but have little in common with photographs of people, and not much in common with one another. Consider too  Figure~\ref{fig:eggy} in which the same parts form a face, or not, depending only on the spatial arrangement of the parts. Indeed, artwork from prehistory to the present day, whether produced by a professional or a child, no matter where in the world: the greater majority of it relies on spatial organisation for recognition.

\section{Applications}

\begin{figure*}[t!]
\centerline{
\includegraphics[width=0.95\textwidth]{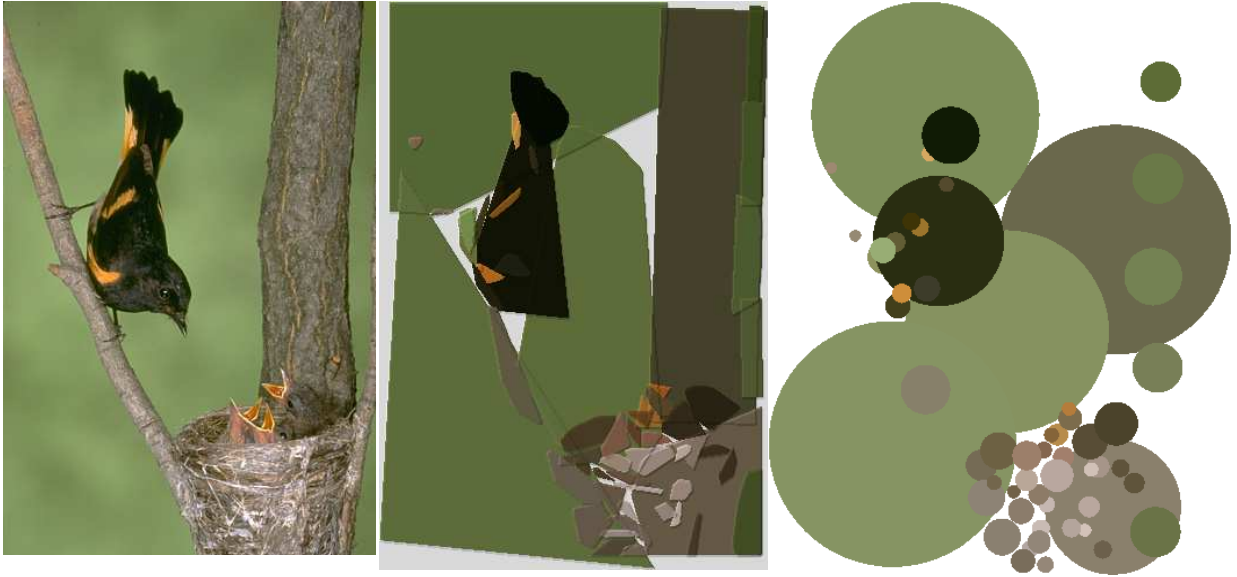}
}
\caption{Shape abstraction for Automated Art.}
\label{fig:shapes}
\end{figure*}

We have already stated that a solution to the cross depiction problem should support advanced application such as web search. It will also support applications such as advanced image editing, examples of which we provide in this section.

Structure, spatial layout, and shape are all important characteristics in identifying objects. These same characteristics can also be used to generate artwork directly from photographs. Consider Figure~\ref{fig:shapes}; it shows a photograph of a bird feeding its young. The photograph has been segmented, and the segments classified into one of a few qualitative shapes (square, circle, triangle, ...). In the most extreme case just one class (circle) is used. See~\cite{song2013abstract} for details of the computer graphics algorithm.

It is true that as the degree of abstraction grows the original interpretation of the image becomes harder to maintain; but given too the degree of abstraction in childrens' drawings, the conclusion that both the quality and quantity of abstraction is important for recognition. In this case the aim was only to produce a ``pretty'' image that bears some resemble to the original. However, simple qualitative shapes of the kind used here can be learned directly from segmentations, as are sufficient to classify scene type (indoor, outdoor, city ...) at close to state-of-the-art rates~\cite{wu2012prime}.

\begin{figure*}[t!]
\centerline{
\includegraphics[width=0.95\textwidth]{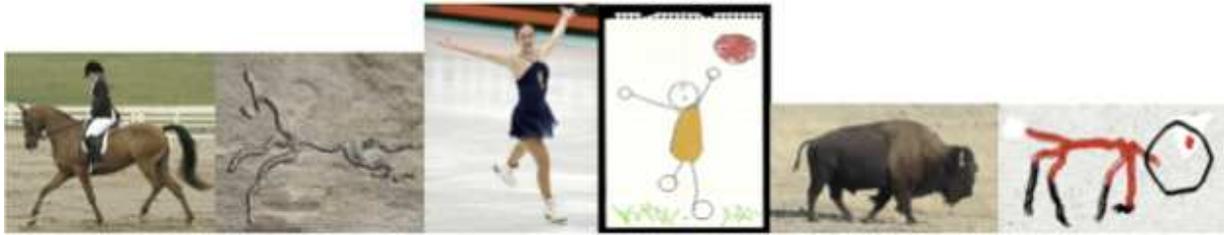}
}
\caption{Structure and Shape combine to make art in the style of (left to right) petroglyphs, child art, Joan Miro.}
\label{fig:art}
\end{figure*}

Shape is not the only form of abstraction useful to the production of art, structure can be used too. Figure~\ref{fig:art} shows examples of computer generated art based on rendering structure. In this case the arcs of a graph have been visualised in a non-photorealistic manner, and the shape of parts at nodes have been classified into a qualitative shape; see~\cite{hall2013simple} for details. An almost identical representation has been used for objects class recognition~\cite{wu-bmvc13}. Even though a lower rates than reported above (around the mid $60\%$ mark) the representation does not exhibit the ``fall off'' when trained on one domain and tested on another, as we have seen with all but one of the methods we have tested in this paper.

\section{Conclusion}
\label{sec:conc}

The cross depiction problem poses an important open problem for Computer Vision. Seeing, as understood outside the field, usually implies parsing a visual signal into semantic objects (I can see {\em it}); in particular it makes no distinction between how those objects are depicted. Our results show that recognition algorithms premised directly on {\em appearance} suffer a fall in performance within the cross-depiction problem; probably because they tacitly assume limited variance of low-level statistics.

All the results we have suggest that spatial organisation between parts is significant with regard to object recognition. For example, DPM out-performs HOG-BoW, even though both use the same low level features; the M-Graph -- with a stronger spatial model -- out-performs DPM. This is because, possibly, structure and spatial layout capture the essential form of an object class, with specific appearance relegated to the level of detail. In other words, structure and space are more salient to robust identification that appearance. Indeed all algorithms we have tested show a significant fall compared to their own peak in performance, when trained on photographs and tested on art; this includes the deep learning methods we have used. The single exception is (\cite{wu2014learning}), which explicitly models a strong structure, and explains appearance details using multiple labels on each node (multiple labels to account for both art and photographic appearance).

Deep learning performs very well on classification over Photo-Art-50, but when presented with the problem of people detection it suffers a significant drop in performance. These equivocal results make it difficult to conclude that deep learning is a solution to the cross-depiction problem; more exactly, the deep learning methods we have tested do not solve cross-depiction.  An alternative network may perform better. 

The cross-depiction problem pushes at the foundations of computer vision, and by doing so it enables new applications; potentially in search, certainly in rendering. Given the fact that the same kind of representations are used both for abstract rendering and for recognition, the conclusion that there is a strong relation between the two is hard to escape -- made more difficult by the observation that people draw what they know not what they see. (When draughting in Art was considered important, students in art school had to be trained to draw what they see rather than what they know.)

In summary: the cross-depiction problem pushes the envelope of computer vision research. It offers significant challenges, which if solved will support a range of applications. Modelling visual classes using structure and spatial relations seems to offer a useful way forward; the role of deep learning in the problem is yet to be fully proven.

\bibliographystyle{plain}
\bibliography{my-ref}

\begin{thebibliography}{10}\itemsep=-1pt

\bibitem{amit2007pop}
Y.~Amit and A.~Trouv{\'e}.
\newblock Pop: Patchwork of parts models for object recognition.
\newblock {\em International Journal of Computer Vision}, 75(2):267--282, 2007.

\bibitem{xiao2011learning}
X.~Bai, Y.-Z. Song, and P.~Hall.
\newblock Learning invariant structure for object identification by using graph
  methods.
\newblock {\em Computer Vision and Image Understanding}, 115(7):1023--1031,
  2011.

\bibitem{yourpaintings}
BBC.
\newblock Your paintings dataset.
\newblock http://www.bbc.co.uk/arts/yourpaintings/.

\bibitem{berg-cvpr01}
A.~C. Berg and J.~Malik.
\newblock Geometric blur for template matching.
\newblock In {\em IEEE International Conference on Computer Vision and Pattern
  Recognition}, 2001.

\bibitem{chatfield-iccv09}
K.~Chatfield, J.~Philbin, and A.~Zisserman.
\newblock Efficient retrieval of deformable shape classes using local
  self-similarities.
\newblock In {\em Workshop on Non-rigid Shape Analysis and Deformable Image
  Alignment, ICCV}, 2009.

\bibitem{Cho_etal_2013}
M.~Cho, K.~Alahari, and J.~Ponce.
\newblock Learning graphs to match.
\newblock In {\em ICCV}.

\bibitem{collomosse2009storyboard}
J.~P. Collomosse, G.~McNeill, and Y.~Qian.
\newblock Storyboard sketches for content based video retrieval.
\newblock In {\em Computer Vision, 2009 IEEE 12th International Conference on},
  pages 245--252. IEEE, 2009.

\bibitem{cootes2001active}
T.~F. Cootes, G.~J. Edwards, and C.~J. Taylor.
\newblock Active appearance models.
\newblock {\em IEEE Transactions on pattern analysis and machine intelligence},
  23(6):681--685, 2001.

\bibitem{coughlan2000efficient}
J.~Coughlan, A.~Yuille, C.~English, and D.~Snow.
\newblock Efficient deformable template detection and localization without user
  initialization.
\newblock {\em Computer Vision and Image Understanding}, 78(3):303--319, 2000.

\bibitem{crandall2005spatial}
D.~Crandall, P.~Felzenszwalb, and D.~Huttenlocher.
\newblock Spatial priors for part-based recognition using statistical models.
\newblock In {\em Computer Vision and Pattern Recognition, 2005. CVPR 2005.
  IEEE Computer Society Conference on}, volume~1, pages 10--17. IEEE, 2005.

\bibitem{CrowleyZisserman_eccvvisart2014}
E.~Crowley and A.~Zisserman.
\newblock In search of art.
\newblock In {\em ECCV Workshop: VisArt}, 2014.

\bibitem{CrowleyZisserman_bmvc2014}
E.~Crowley and A.~Zisserman.
\newblock The state of the art: Object retrieval in paintings using
  discriminative regions.
\newblock In {\em Proceedings of the British Machine Vision Conference}. BMVA
  Press, 2014.

\bibitem{crowley2013gods}
E.~J. Crowley and A.~Zisserman.
\newblock Of gods and goats: Weakly supervised learning of figurative art.
\newblock {\em learning}, 8:14, 2013.

\bibitem{csurka-eccv04}
G.~Csurka, C.~R. Dance, L.~Fan, J.~Willamowski, and C.~Bray.
\newblock Visual categorization with bags of keypoints.
\newblock In {\em Workshop on Statistical Learning in Computer Vision, ECCV},
  pages 1--22, 2004.

\bibitem{dalal-cvpr05}
N.~Dalal and B.~Triggs.
\newblock Histograms of oriented gradients for human detection.
\newblock In {\em CVPR}, volume~2, pages 886--893, 2005.

\bibitem{eitz-tvcg11}
M.~Eitz, K.~Hildebrand, T.~Boubekeur, and M.~Alexa.
\newblock Sketch-based image retrieval: Benchmark and bag-of-features
  descriptors.
\newblock {\em IEEE Transactions on Visualization and Computer Graphics},
  17(11):1624--1636, 2011.

\bibitem{eitz-siggraph12}
M.~Eitz, R.~Richter, T.~Boubekeur, K.~Hildebrand, and M.~Alexa.
\newblock Sketch-based shape retrieval.
\newblock {\em ACM Trans. Graph. (Proc. SIGGRAPH)}, 31(4):31:1--31:10, 2012.

\bibitem{elidan2006learning}
G.~Elidan, G.~Heitz, and D.~Koller.
\newblock Learning object shape: From drawings to images.
\newblock In {\em Computer Vision and Pattern Recognition, 2006 IEEE Computer
  Society Conference on}, volume~2, pages 2064--2071. IEEE, 2006.

\bibitem{everingham2010pascal}
M.~Everingham, L.~Van~Gool, C.~K. Williams, J.~Winn, and A.~Zisserman.
\newblock The pascal visual object classes (voc) challenge.
\newblock {\em International journal of computer vision}, 88(2):303--338, 2010.

\bibitem{felzenszwalb2010object}
P.~F. Felzenszwalb, R.~B. Girshick, D.~McAllester, and D.~Ramanan.
\newblock Object detection with discriminatively trained part-based models.
\newblock {\em Pattern Analysis and Machine Intelligence, IEEE Transactions
  on}, 32(9):1627--1645, 2010.

\bibitem{felzenszwalb-pami10}
P.~F. Felzenszwalb, R.~B. Girshick, D.~McAllester, and D.~Ramanan.
\newblock Object detection with discriminatively trained part based models.
\newblock {\em IEEE Transactions on Pattern Analysis and Machine Intelligence},
  32(9):1627--1645, 2010.

\bibitem{felzenszwalb2005pictorial}
P.~F. Felzenszwalb and D.~P. Huttenlocher.
\newblock Pictorial structures for object recognition.
\newblock {\em International Journal of Computer Vision}, 61(1):55--79, 2005.

\bibitem{fergus2003object}
R.~Fergus, P.~Perona, and A.~Zisserman.
\newblock Object class recognition by unsupervised scale-invariant learning.
\newblock In {\em Computer Vision and Pattern Recognition, 2003. Proceedings.
  2003 IEEE Computer Society Conference on}, volume~2, pages II--264. IEEE,
  2003.

\bibitem{fernando-iccv13}
B.~Fernando, A.~Habrard, M.~Sebban, and T.~Tuytelaars.
\newblock Unsupervised visual domain adaptation using subspace alignment.
\newblock In {\em ICCV}, 2013.

\bibitem{ferrari2008groups}
V.~Ferrari, L.~Fevrier, F.~Jurie, and C.~Schmid.
\newblock Groups of adjacent contour segments for object detection.
\newblock {\em Pattern Analysis and Machine Intelligence, IEEE Transactions
  on}, 30(1):36--51, 2008.

\bibitem{eth-shape}
V.~Ferrari, F.~Jurie, and C.~Schmid.
\newblock From images to shape models for object detection.
\newblock {\em IJCV}, 2010.

\bibitem{fischler1973representation}
M.~A. Fischler and R.~A. Elschlager.
\newblock The representation and matching of pictorial structures.
\newblock {\em IEEE Transactions on Computers}, 22(1):67--92, 1973.

\bibitem{Ginosar_etal_eccvvisart2014}
S.~Ginosar, D.~Haas, T.~Brown, and J.~Malik.
\newblock Detecting people in cubist art.
\newblock In {\em ECCV Workshop: VisArt}, 2014.

\bibitem{girshick-cvpr14}
R.~Girshick, J.~Donahue, T.~Darrell, and J.~Malik.
\newblock Rich feature hierarchies for accurate object detection and semantic
  segmentation.
\newblock In {\em Proceedings of the IEEE Conference on Computer Vision and
  Pattern Recognition ({CVPR})}, 2014.

\bibitem{gong-icml13}
B.~Gong, K.~Grauman, and F.~Sha.
\newblock Connecting the dots with landmarks: Discriminatively learning
  domain-invariant features for unsupervised domain adaptation.
\newblock In {\em ICML}, pages 222--230, 2013.

\bibitem{gong-cvpr12}
B.~Gong, Y.~Shi, F.~Sha, and K.~Grauman.
\newblock Geodesic flow kernel for unsupervised domain adaptation.
\newblock In {\em CVPR}, pages 2066--2073, 2012.

\bibitem{gopalan-iccv11}
R.~Gopalan, R.~Li, and R.~Chellappa.
\newblock Domain adaptation for object recognition: An unsupervised approach.
\newblock In {\em IEEE International Conference on Computer Vision}, volume~0,
  pages 999--1006, 2011.

\bibitem{gu2009recognition}
C.~Gu, J.~J. Lim, P.~Arbelaez, and J.~Malik.
\newblock Recognition using regions.
\newblock In {\em Computer Vision and Pattern Recognition, 2009. CVPR 2009.
  IEEE Conference on}, pages 1030--1037. IEEE, 2009.

\bibitem{hall2013simple}
P.~Hall and Y.-Z. Song.
\newblock Simple art as abstractions of photographs.
\newblock In {\em Proceedings of the Symposium on Computational Aesthetics},
  pages 77--85. ACM, 2013.

\bibitem{hu-icip10}
R.~Hu, M.~Barnard, and J.~P. Collomosse.
\newblock Gradient field descriptor for sketch based retrieval and
  localization.
\newblock In {\em ICIP}, pages 1025--1028, 2010.

\bibitem{hu-cviu13}
R.~Hu and J.~Collomosse.
\newblock A performance evaluation of gradient field hog descriptor for sketch
  based image retrieval.
\newblock {\em Computer Vision and Image Understanding}, 117(7):790--806, 2013.

\bibitem{hu2013markov}
R.~Hu, S.~James, T.~Wang, and J.~Collomosse.
\newblock Markov random fields for sketch based video retrieval.
\newblock In {\em Proceedings of the 3rd ACM conference on International
  conference on multimedia retrieval}, pages 279--286. ACM, 2013.

\bibitem{jia2010classifying}
W.~Jia and S.~J. McKenna.
\newblock Classifying textile designs using bags of shapes.
\newblock In {\em ICPR}, pages 294--297, 2010.

\bibitem{jia-caffe14}
Y.~Jia, E.~Shelhamer, J.~Donahue, S.~Karayev, J.~Long, R.~Girshick,
  S.~Guadarrama, and T.~Darrell.
\newblock Caffe: Convolutional architecture for fast feature embedding.
\newblock {\em arXiv preprint arXiv:1408.5093}, 2014.

\bibitem{krizhevsky-nips12}
A.~Krizhevsky, I.~Sutskever, and G.~E. Hinton.
\newblock Imagenet classification with deep convolutional neural networks.
\newblock In F.~Pereira, C.~Burges, L.~Bottou, and K.~Weinberger, editors, {\em
  Advances in Neural Information Processing Systems 25}, pages 1097--1105.
  Curran Associates, Inc., 2012.

\bibitem{lazebnik-cvpr06}
S.~Lazebnik, C.~Schmid, and J.~Ponce.
\newblock Beyond bags of features: Spatial pyramid matching for recognizing
  natural scene categories.
\newblock In {\em CVPR}, volume~2, pages 2169--2178. IEEE, 2006.

\bibitem{leibe2008robust}
B.~Leibe, A.~Leonardis, and B.~Schiele.
\newblock Robust object detection with interleaved categorization and
  segmentation.
\newblock {\em International journal of computer vision}, 77(1-3):259--289,
  2008.

\bibitem{Leordeanu_etal_cvpr2007}
M.~Leordeanu and M.~H. andR. Sukthankar.
\newblock Beyond local appearance: Category recognition from pairwise
  interactions of simple features.
\newblock In {\em CVPR}, 2007.

\bibitem{li2013sketch}
Y.~Li, Y.-Z. Song, and S.~Gong.
\newblock Sketch recognition by ensemble matching of structured features.
\newblock In {\em In British Machine Vision Conference (BMVC)}. Citeseer, 2013.

\bibitem{lowe-ijcv04}
D.~G. Lowe.
\newblock Distinctive image features from scale-invariant keypoints.
\newblock {\em Intl.\ Journal of Computer Vision}, 60(2):91--110, 2004.

\bibitem{pan-nn11}
S.~J. Pan, I.~W. Tsang, J.~T. Kwok, and Q.~Y. 0001.
\newblock Domain adaptation via transfer component analysis.
\newblock {\em IEEE Transactions on Neural Networks}, 22(2):199--210, 2011.

\bibitem{perronnin-eccv10}
F.~Perronnin, J.~S\'{a}nchez, and T.~Mensink.
\newblock Improving the fisher kernel for large-scale image classification.
\newblock In {\em Proceedings of the 11th European Conference on Computer
  Vision: Part IV}, ECCV'10, pages 143--156, 2010.

\bibitem{rom1993hierarchical}
H.~Rom and G.~Medioni.
\newblock Hierarchical decomposition and axial shape description.
\newblock {\em Pattern Analysis and Machine Intelligence, IEEE Transactions
  on}, 15(10):973--981, 1993.

\bibitem{russakovsky-eccv12}
O.~Russakovsky, Y.~Lin, K.~Yu, and L.~Fei-Fei.
\newblock Object-centric spatial pooling for image classification.
\newblock In {\em ECCV}, 2012.

\bibitem{saenko-eccv10}
K.~Saenko, B.~Kulis, M.~Fritz, and T.~Darrell.
\newblock Adapting visual category models to new domains.
\newblock In {\em European Conference on Computer Vision}, pages 213--226,
  2010.

\bibitem{shechtman2007matching}
E.~Shechtman and M.~Irani.
\newblock Matching local self-similarities across images and videos.
\newblock In {\em Computer Vision and Pattern Recognition, 2007. CVPR'07. IEEE
  Conference on}, pages 1--8. IEEE, 2007.

\bibitem{shrivastava-tog11}
A.~Shrivastava, T.~Malisiewicz, A.~Gupta, and A.~A. Efros.
\newblock Data-driven visual similarity for cross-domain image matching.
\newblock {\em ACM Transaction of Graphics (TOG) (Proceedings of ACM SIGGRAPH
  ASIA)}, 30(6), 2011.

\bibitem{siddiqi1999shock}
K.~Siddiqi, A.~Shokoufandeh, S.~J. Dickinson, and S.~W. Zucker.
\newblock Shock graphs and shape matching.
\newblock {\em International Journal of Computer Vision}, 35(1):13--32, 1999.

\bibitem{song2013abstract}
Y.-Z. Song, D.~Pickup, C.~Li, P.~Rosin, and P.~Hall.
\newblock Abstract art by shape classification.
\newblock {\em Visualization and Computer Graphics, IEEE Transactions on},
  19(8):1252--1263, 2013.

\bibitem{sundar2003skeleton}
H.~Sundar, D.~Silver, N.~Gagvani, and S.~Dickinson.
\newblock Skeleton based shape matching and retrieval.
\newblock In {\em Shape Modeling International, 2003}, pages 130--139. IEEE,
  2003.

\bibitem{uijlings-ijcv13}
J.~Uijlings, K.~van~de Sande, T.~Gevers, and A.~Smeulders.
\newblock Selective search for object recognition.
\newblock {\em International Journal of Computer Vision}, 2013.

\bibitem{vlfeat}
A.~Vedaldi and B.~Fulkerson.
\newblock {VLFeat}: An open and portable library of computer vision algorithms.
\newblock http://www.vlfeat.org/, 2008.

\bibitem{vedaldi-cvpr10}
A.~Vedaldi and A.~Zisserman.
\newblock Efficient additive kernels via explicit feature maps.
\newblock In {\em IEEE Conf. on Computer Vision and Pattern Recognition
  (CVPR)}, 2010.

\bibitem{wu2014learning}
Q.~Wu, H.~Cai, and P.~Hall.
\newblock Learning graphs to model visual objects across different depictive
  styles.
\newblock In {\em Computer Vision--ECCV 2014}, pages 313--328. Springer, 2014.

\bibitem{wu2012prime}
Q.~Wu and P.~Hall.
\newblock Prime shapes in natural images.
\newblock In {\em BMVC}, pages 1--12, 2012.

\bibitem{wu-bmvc13}
Q.~Wu and P.~Hall.
\newblock Modelling visual objects invariant to depictive style.
\newblock In {\em BMVC}, 2013.

\bibitem{yang2011articulated}
Y.~Yang and D.~Ramanan.
\newblock Articulated pose estimation with flexible mixtures-of-parts.
\newblock In {\em Computer Vision and Pattern Recognition (CVPR), 2011 IEEE
  Conference on}, pages 1385--1392. IEEE, 2011.

\end{thebibliography}

\end{document}